\documentclass[lettersize,journal]{IEEEtran}
\usepackage{amsmath,amsfonts}
\usepackage{algorithmic}
\usepackage{algorithm}
\usepackage{array}
\usepackage[caption=false,font=normalsize,labelfont=sf,textfont=sf]{subfig}
\usepackage{textcomp}
\usepackage{stfloats}
\usepackage{url}
\usepackage{verbatim}
\usepackage{graphicx}
\usepackage{cite}
\usepackage{amssymb}
\newcommand{\vect}[1]{\mathbf{#1}}

\newcommand{\etal}{\textit{et al.}}
\usepackage{booktabs}
\usepackage{multirow}
\usepackage{listings}
\usepackage{soul, color}
\usepackage{hyperref}
\begin{document}

\title{Dark-EvGS: Event Camera as an Eye for Radiance Field in the Dark}

\author{Jingqian Wu, Peiqi Duan\textsuperscript{\dag}, Zongqiang Wang, Changwei Wang, \\ Boxin Shi,~\IEEEmembership{Senior Member, IEEE} and Edmund Y. Lam\textsuperscript{\dag},~\IEEEmembership{Fellow, IEEE}
\thanks{Jingqian Wu and Edmund Y. Lam are with the Department of Electrical and Electronic Engineering, The University of Hong Kong, Pokfulam, Hong Kong SAR, China.}
\thanks{Peiqi Duan and Boxin Shi are with the State Key Laboratory of Multimedia Information Processing and National Engineering Research Center of Visual Technology, School of Computer Science, Peking University, Beijing, China.}
\thanks{Zongqiang Wang is with the Institute of Automation, Chinese Academy of Sciences, Beijing, China.}
\thanks{Changwei Wang is with the Key Laboratory of Computing Power Network and Information Security, Ministry of Education, Shandong Computer Science Center, Qilu University of Technology, Shandong, China.}
\thanks{\textsuperscript{\dag} Corresponding authors: duanqi0001@pku.edu.cn; elam@eee.hku.hk}
}

\markboth{IEEE TRANSACTIONS ON IMAGE PROCESSING}%
{Shell \MakeLowercase{\textit{et al.}}: A Sample Article Using IEEEtran.cls for IEEE Journals}


\maketitle

\begin{abstract}
In low-light environments, conventional cameras often struggle to capture clear multi-view images of objects due to dynamic range limitations and motion blur caused by long exposure. Event cameras, with their high-dynamic range and high-speed properties, have the potential to mitigate these issues. Additionally, 3D Gaussian Splatting (GS) enables radiance field reconstruction, facilitating bright frame synthesis from multiple viewpoints in low-light conditions. However, naively using an event-assisted 3D GS approach still faced challenges because, in low lights, events are noisy, frames lack quality, and the color tone may be inconsistent. To address these issues, we propose Dark-EvGS, the first event-assisted 3D GS framework that enables the reconstruction of bright frames from arbitrary viewpoints along the camera trajectory. Triplet-level supervision is proposed to gain holistic knowledge, granular details, and sharp scene rendering. The color tone matching block is proposed to guarantee the color consistency of the rendered frames. Furthermore, we introduce the first real-captured dataset for the event-guided bright frame synthesis task via 3D GS-based radiance field reconstruction. Experiments demonstrate that our method achieves better results than existing methods, conquering radiance field reconstruction under challenging low-light conditions. The code and sample data are included in the supplementary material.
\end{abstract}


\section{Introduction}
Low-light radiance field reconstruction plays a crucial role in various real-world applications, including nighttime photography \cite{liu2024ntire, zhu2025modeling, lyu2024enhancing}, surveillance \cite{liu2024seeing}, and autonomous driving \cite{gehrig2024low, wu2025segment, lu2025see}. However, traditional frame-based cameras struggle to capture high-quality images in dark environments due to their limited dynamic range and reliance on long exposure times. These constraints result in significant noise and motion blur, making it difficult to capture clear, bright frames of objects from multiple viewpoints in low-light conditions.

Event cameras are a novel type of sensor that captures brightness changes asynchronously per pixel, marking a shift from conventional frame-based imaging \cite{yang2024latency, ding2022spatio, chen2022progressivemotionseg, ge2025event, liu2024line}. With high dynamic range (HDR) and high temporal resolution capabilities, they can simultaneously capture bright and dark regions with no motion blur, enabling the recovery of visual details often lost in frame-based imaging, and even supporting radiance field reconstruction, particularly under low-light conditions. Event-based video reconstruction methods are proposed \cite{rebecq2019events, ercan2024hypere2vid, liang2023coherent} to output consecutive frames from events. Nevertheless, either they are limited to grayscale images \cite{rebecq2019events, ercan2024hypere2vid}, or primarily focus on the short exposure problem \cite{liang2023coherent}. EvLight \cite{liang2024towards} leverages event data for single-image enhancement rather than focusing on video, and may exhibit partial chromatic aberrations. Event data have also been applied to radiance field reconstruction, but existing approaches \cite{wu2024EV-GS, rudnev2023eventnerf, xiong2024event3dgs} are only designed for normal light conditions (around 300 lux). 

\begin{figure}[t]
	
	\centering
	
	\includegraphics[width=\linewidth,scale=1.0]{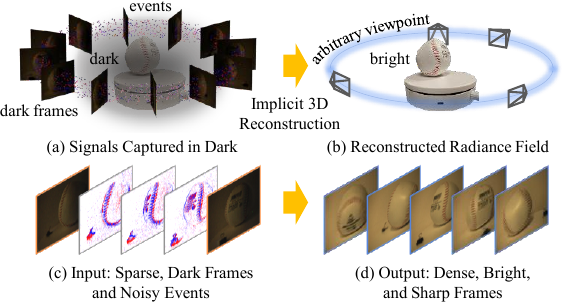}
	
	\caption{In the dark, only noisy events and dark blurred frames can be captured (a), which is challenging for existing approaches for radiance field reconstruction. Our approach takes the raw events and sparse dark frames as input (c), and it reconstructs a radiance field, enabling arbitrary viewpoint synthesis (b). Our approach is capable of synthesizing dense, bright, and sharp views, enhancing visibility and detail (d).}
	
	\label{fig:Intro}
	
\end{figure}

Reconstructing a radiance field in the dark is particularly challenging due to several factors. 
First, the captured signals lack quality: dark frames contain extremely low intensity and brightness, while event streams become significantly noisier and sparser because of photon scarcity (Fig.~\ref{fig:Intro} (a) and (c)). This makes both frame-only methods (e.g., 3D GS~\cite{kerbl20233d}) and event-based radiance field approaches~\cite{wu2024EV-GS, rudnev2023eventnerf, xiong2024event3dgs} ineffective under such conditions. 
Second, ensuring color tone consistency across rendered views is non-trivial. Existing event-based video reconstruction works \cite{rebecq2019events, ercan2024hypere2vid, liang2023coherent} either output grayscale sequences or focus on short-exposure enhancement, and thus cannot be directly adapted for radiance field reconstruction in dark environments. 
Third, accurate camera pose estimation and data availability present additional obstacles. On the one hand, structure-from-motion pipelines struggle in dark environments because frames lack sufficient intensity for reliable feature extraction, and no mature event-based COLMAP alternative exists that can robustly handle noisy, sparse events. On the other hand, the absence of dedicated datasets further limits progress: most prior studies rely on synthetic short/long exposure pairs and simulated events~\cite{hu2021v2e}, which fail to capture the noise characteristics, dynamic lighting, and inconsistencies of real-world low-light data. As a result, even when accurate ground-truth poses are provided in controlled experiments, such settings do not reflect real-world deployments, and models trained on synthetic data generalize poorly to real-world low light datasets and environments.

In this paper, we propose Dark-EvGS by integrating the event camera and 3D Gaussian Splatting (GS) \cite{kerbl20233d}, a radiance field reconstruction technique, enabling bright frame reconstruction from arbitrary viewpoints along the camera trajectory. The high-dynamic sensing capability of event cameras enables Dark-EvGS to have a precise perception of camera motion, which can provide valuable guidance for radiance field reconstruction performance (Fig. \ref{fig:Intro} (b)) and render bright frames from multiple viewpoints of objects in low-light conditions (Fig. \ref{fig:Intro} (d)).

Dark-EvGS conquers the mentioned challenges in three directions: 1) We propose a triplet-level supervision mechanism specifically tailored for handling noisy event and frame signals in low-light radiance field reconstruction; 2) the Color Tone Matching Block (CTMB) is proposed to guarantee color consistency of the rendered frames; 3) we collect a dataset for the event-guided bright frame synthesis task via 3D GS-based radiance field reconstruction, which includes paired low-light frames, bright ground truth frames, event streams, and corresponding camera poses. Extensive experiments demonstrate that our method outperforms previous state-of-the-art techniques across all samples. The outlined key technical contributions are: 
\begin{itemize} 
\item We introduce Dark-EvGS, the first event-guided bright frame synthesis method via 3D GS-based radiance field reconstruction from arbitrary viewpoints in the dark.

\item We present a triplet-level supervision strategy to recover missed holistic structures, restore fine-grained details, and enhance sharpness for 3D GS training and for rendering color-consistent bright frames in low-light environments.

\item We collect the first real-world event-based low-light radiance field reconstruction dataset with paired low-light, bright-light frames, event streams, and corresponding camera poses. We will release the full dataset and code to facilitate future research. 
\end{itemize}

\section{Related Work}
\label{sec:related}

\noindent\textbf{General and Dark Radiance Field.}
Radiance field rendering and novel-view synthesis are fundamental tasks in graphics and computer vision with extensive applications in fields like robotics and virtual reality. Neural Radiance Fields (NeRF) \cite{mildenhall2021nerf} and 3D GS \cite{kerbl20233d} have made substantial progress in these tasks. While NeRF produces high-fidelity renderings with intricate detail, the high number of samples required to accumulate the color information for each pixel results in low rendering efficiency and extended training times \cite{kerbl20233d}. 3D GS, on the other hand, employs a set of optimized Gaussians to achieve state-of-the-art quality in reconstruction and rendering speed \cite{wu2024EV-GS}. Starting from sparse point clouds generated by Structure-from-Motion (SfM), 3D GS uses differentiable rendering to control the density and refine parameters adaptively.

Most NeRF and 3D GS-based methods use frame data from traditional cameras \cite{mildenhall2022nerf, kerbl20233d}. Few works have attempted to reconstruct radiance fields in dark and low-light environments due to the limited dynamic range of a frame-based camera. Zhang \etal \cite{zhang2024darkgs} attempt to solve robot exploration in the dark using 3D Gaussians Relighting, but a constant illumination placed in front of the robot is necessary. Mildenhall \etal \cite{mildenhall2022nerf} use NeRF to reconstruct the radiance field in the dark by treating it as a denoising problem from HDR frames. However, it requires an HDR sensor in real applications and does not explore the effect of motion blur in low-light conditions. Some works \cite{huang2025inceventgs, wu2025sweepevgs, wu2024EV-GS, feng20254dgs} have devoted to event-based radiance field reconstruction. But they all reconstruct under normal lighting conditions. Thus, limited work has been devoted to leveraging the dynamic range and temporal resolution properties of an event camera to solve such an issue.

\noindent\textbf{Event-Based Video Reconstruction.}
Many works have been devoted to the field of events to video reconstruction \cite{rebecq2019events, qu2024e2hqv, ercan2024hypere2vid}. However, these methods do not work well for reconstructing frames from low-light events. EvLowLight \cite{liang2023coherent} attempts to reconstruct bright-light video by incorporating event data. However, the focus is primarily on the short-to-long exposure problem rather than the dark-to-bright reconstruction. In their case, the darkness of the frames is caused by short exposure times, not actual low-light conditions, meaning the scene remains well-lit from the event camera's perspective. EvLight \cite{liang2024towards} utilizes event data to enhance single images rather than addressing video reconstruction, and its results may still exhibit partial chromatic aberrations. 
In contrast, our task focuses on using an event camera to reconstruct bright radiance fields in real low-light environments. Others \cite{zhang2020learning, paredes2021back} have tried to use an event to guide video reconstruction in the dark. However, estimating absolute intensity values in a video solely from brightness changes recorded in events is a highly ill-posed problem \cite{liang2023coherent}. Thus, directly applying any of these models to our task is not feasible. 

\noindent\textbf{Event-Based Radiance Field Reconstruction.}
Event cameras, with their unique characteristics, such as motion-blur resistance, high dynamic range, low latency, and energy efficiency, are increasingly utilized in computer vision and computational imaging applications \cite{han2023hybrid, duan2023neurozoom, zhang2024joint, zhu2024efficient}. Approaches integrating NeRF with event data \cite{klenk2023nerf, rudnev2023eventnerf} apply volumetric rendering using event supervision, which takes advantage of NeRF’s inherent multi-view consistency to extract coherent scene structures. However, the computational demands for training and optimizing event-based NeRF pipelines remain substantial \cite{wu2024EV-GS}. Recent works have aimed to integrate 3D GS with event data \cite{wu2024EV-GS, yu2024evagaussians, xiong2024event3dgs, wu2025sweepevgs, zahid20253dgs, deguchi2024e2gs}. Regardless of the base approach, existing event-involved approaches were designed for general scenes. For those who take event streams as input only, the noisier and more random nature of events under dark and low-light environments makes these approaches unrobust and ineffective for reconstruction. For others who also utilize blurred frames as additional input, it further suffers from the low intensity in low-light frames.

\section{Methodology}
\label{sec:method}

\subsection{Preliminary on 3D Gaussian Splatting}
\begin{figure*}[t]
\begin{center}
  \includegraphics[width=1 \linewidth]{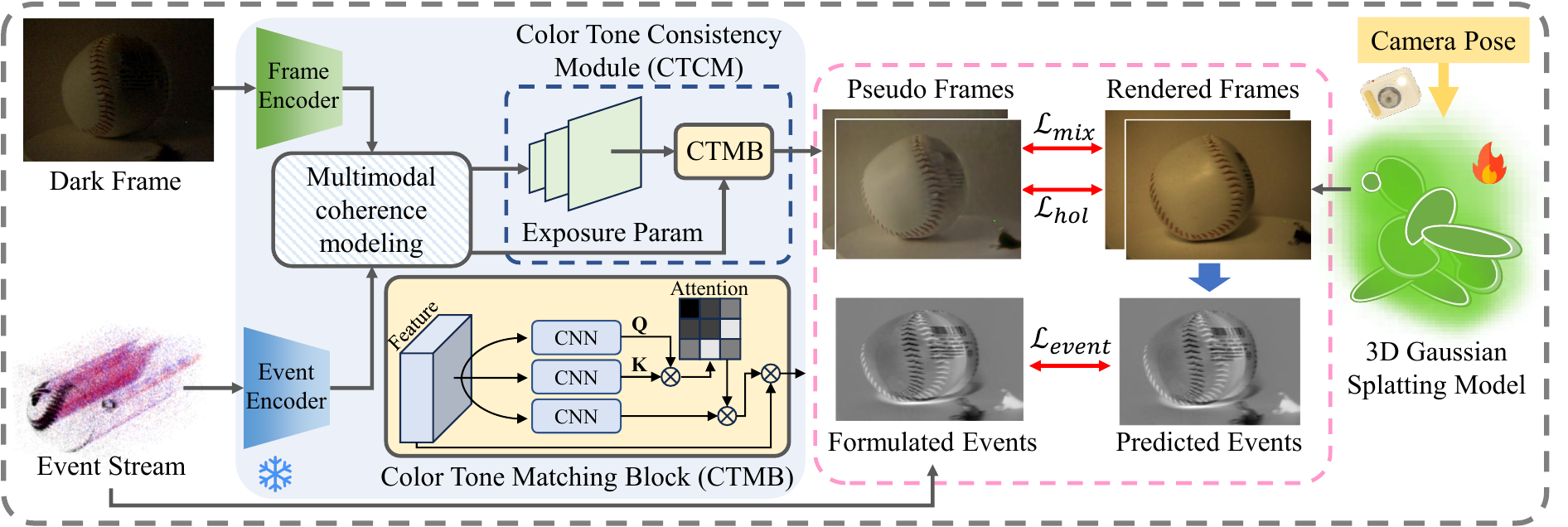}
\end{center}
   \caption{Overview of the Dark-EvGS pipeline for radiance field reconstruction in the dark. We obtain dark frames, events, and camera parameters using an event camera under low light. The frame encoder and event encoder will extract features, which will then be forwarded to multimodal coherence modeling. The proposed Color Tone Consistency Module takes the features and decodes them into pseudo bright frames with consistent color tone (described in Sec. \ref{CTMB}). We used pre-trained weights from \cite{liang2023coherent} for the initialization of encoders and the multimodal coherence modeling block. The ice icon denotes frozen weights, and the fire icon denotes optimizing weights. Via the 3D GS model, we render bright and shape frames when given the corresponding camera position and pose, and formulate (blue arrow) the rendered results to predicted event signals. The same process applies when formulating captured events to ground truth supervision signals. To supervise accurate training, the proposed triplet-level loss (red lines) provides a holistic view while keeping the rendered results sharp and accurate in detail. As a result, our method enables high-quality radiance field reconstruction in the dark.}
\label{fig:pipline}
\end{figure*}

\label{Sec: 3D GS}
The 3D Gaussian Splatting (3D GS) approach \cite{kerbl20233d} models detailed 3D scenes as point clouds, where each point is represented as a Gaussian, defining the structure of the scene. Each Gaussian is characterized by a 3D covariance matrix $\Sigma$ and a central location $\vect{x}$:

\begin{equation} G(x) = \exp \left(-\frac{1}{2}\vect{x}^{T}\Sigma^{-1}\vect{x} \right), \label{GS_eq1} \end{equation}
where the central location $\vect{x}$ serves as the Gaussian's mean. To enable differentiable optimization, $\Sigma$ is decomposed into a rotation matrix $R$ and a scaling matrix $S$:

\begin{equation} \Sigma = RSS^TR^T. \end{equation}

Rendering from various viewpoints utilizes the splatting technique from \cite{yifan2019differentiable}, which projects Gaussians onto camera planes. This technique, based on the method in \cite{zwicker2001surface}, involves the viewing transformation matrix $W$ and the Jacobian $J$ of the affine projective transformation. In-camera coordinates, the covariance matrix $\Sigma'$ is then represented as:

\begin{equation} \Sigma' = J W ~\Sigma ~W^{T}J^{T}. \label{Eq: GS} \end{equation}

To summarize, each Gaussian point includes several attributes: position $\vect{x} \in \mathbb{R}^3$, color encoded via spherical harmonics coefficients $\vect{c} \in \mathbb{R}^k$ (with $k$ indicating the degrees of freedom), opacity $\alpha \in \mathbb{R}$, rotation quaternion $\vect{q} \in \mathbb{R}^4$, and scaling factor $\vect{s} \in \mathbb{R}^3$. For each pixel, the combined color and opacity values of all Gaussians are computed using the Gaussian representation from Equation \ref{GS_eq1}. The blending of colors $C$ for $N$ ordered points projected onto a pixel is:

\begin{equation} C = \sum_{i \in N} \vect{c}_i \alpha_i \prod_{j=1}^{i-1} (1-\alpha_j), \end{equation}
where $\vect{c}_i$ and $\alpha_i$ represent the color and density of each point, respectively. These values are influenced by each Gaussian's covariance matrix $\Sigma$ and modified through per-point opacity and spherical harmonics color coefficients.

\subsection{Noisy Events in Dark as Supervisory Signals} 
\label{Sec: Event Stream Utilization} 



The event sensor output can be formulated as:

\begin{equation}
    E_{gt} = \Gamma \left\{ \log \left( \frac{I_{t} + c}{I_{t-w} + c} \right), \epsilon \right\},
\label{Eq: event_formulation}
\end{equation}

where $E_{gt}$ is the captured events, $I_t$ and $I_{t-w}$ are intensity images at the two timestamps, $\Gamma\{\theta, \epsilon\}$ represents the conversion function from log-intensity to events, $c$ is an offset value to prevent $\log(0)$, $\epsilon$ is the event triggering threshold. $\Gamma\{\theta, \epsilon\} = 1$ when $\theta \geq \epsilon$, indicating positive event triggered, and $\Gamma\{\theta, \epsilon\} = -1$ when $\theta \leq -\epsilon$, indicating negative event triggered \cite{duan2021eventzoom}. Else, no events are triggered.

\begin{figure}[t]
	
	\centering
	
	\includegraphics[width=\linewidth,scale=1.0]{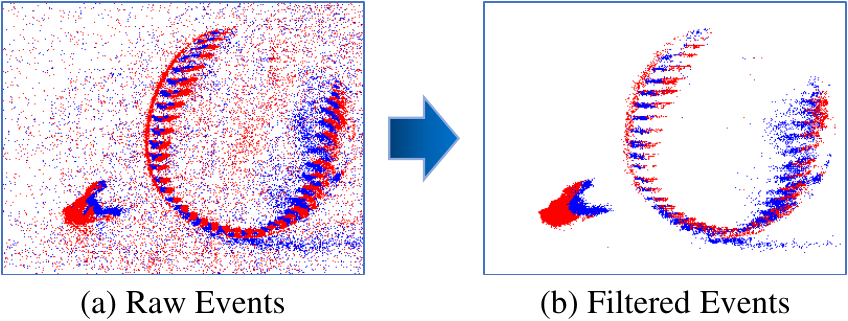}
	
	\caption{Frame-based cameras are unable to capture enough signals for residence reconstruction in the dark. Event cameras, on the other hand, can capture ample signals, including ignored details by frame-based cameras (a). However, these raw events captured are rather noisy and random (a), which is one of the biggest challenges for existing methods \cite{liang2023coherent, wu2024EV-GS, rudnev2023eventnerf, rebecq2019events}. To solve this problem, we proposed a utilization for noisy events for accurate and clean supervisory signals (b).}
	
	\label{fig:noise filter}
	
\end{figure}

Events triggered in low-light environments are much noisier \cite{hu2021v2e, fang2022aednet}. To address noise, which is often more prevalent in microsecond-level event data \cite{yang2023sci}, we preprocess the event stream using the $y$-noise filter from \cite{feng2020event}. As shown in Fig. \ref{fig:noise filter} (a), raw events captured in the dark contain extreme noise and randomness. After applying the noise filter, the cleaned event stream becomes more helpful as a supervision signal, as shown in Fig. \ref{fig:noise filter} (b). Noise filtering is essential for accurate view rendering, as shown by the ablation study section.

Our objective, depicted in Fig. \ref{fig:pipline}, is to reconstruct a radiance field using differentiable 3D Gaussian functions $G$, guided by event and corresponding blurred dark frames. We accumulate the ground truth event signal between timestamps $t_1$ and $t_2$ to form the supervisory signal using Eq. \ref{Eq: event_formulation} by aggregating the polarities of all events occurring between times $t$ and $t-w$, indexed by their pixel location.


At each timestamp, $t_k$, a rendering result is produced with the camera at pose $p_k$. Thus, we calculate two rendered frames from the 3D GS model: $I_1 = G(p_1)$ and $I_2 = G(p_2)$, where $I_1$ and $I_2$ are RGB renderings at times $t_1$ and $t_2$, respectively. Here, $G$ denotes the 3D GS model, and $p_1$ and $p_2$ are the camera poses at timestamps $t_1$ and $t_2$. The logarithmic image representation is defined as $L(I_t) = \log\big((I_t)^g + \kappa\big)$, where $\kappa = 1 \times 10^{-5}$ (to prevent NaN values) and $g$ is a gamma correction factor set to 2.2 across all experiments, following \cite{international1999multimedia, rudnev2023eventnerf}. From this, we obtain the predicted cumulative difference $E_{pred} = L(I_2) - L(I_1)$.

\subsection{Color Tone Consistency Module}
\label{CTMB}
Pure event utilization and supervision are insufficient for radiance field reconstruction on real data \cite{han2025event, wu2024EV-GS}, let alone reconstruction under low-light environments. Therefore, prior knowledge is required in the 3D GS training pipeline. Following EvLowLight \cite{liang2023coherent}, we deploy feature encoders for both frame and event modality along with the Multimodal Coherence Modeling block (MCM), with corresponding pre-trained weights for initialization. We then propose the Color Tone Consistency Module (CTCM), which consists of a proposed Color Tone Matching Block (CTMB) along with the decoders from MCM. CTMB takes the decoded frames during the decoding process and corrects the color tone. Finally, the CTCM outputs the predicted pseudo frames using the corrected frames and the exposure parameter predicted from the MCM. Specifically, the CTMB, inspired by recent advancements in transposed self-attention \cite{vaswani2017attention}, takes the last feature map \( F \in \mathbb{R}^{H \times W \times C} \), and derives query \( Q \), key \( K \), and value \( V \) representations using a 1×1 convolution followed by a depth-wise convolution. Subsequently, \( Q \) and \( V \) are reshaped, and an attention map \( M \in \mathbb{R}^{C \times C} \) is computed through matrix multiplication. The output of the CTMB \( F_{\text{out}} \) represents the color-corrected feature, and the bright light frame with corrected color tone can be derived from \( F_{\text{out}} \) and the exposure parameter from MCM. This allows CTCM to generate global color correction via channel-wise self-attention while enhancing local color adjustments through CNNs. 

\subsection{Triplet Supervision for Radiance Field in Dark} 
\label{Sec: Supervision} 
There are three major challenges in supervision under low light conditions: First, events are too noisy to provide accurate supervision alone. Second, pseudo-bright frames generated by CTCM cannot provide robust details due to the data gap between pre-trained knowledge and actual training scenarios. Third, in low light environments, captured dark frames are easier to motion blur as a longer exposure time is required \cite{zhu2025learning, chen2025evlight++}. Thus, a sharpening module is needed because the pseudo-bright frames may also contain motion blur when given blurred dark frames. 

To tackle these challenges, we propose triplet-level supervision to train and reconstruct radiance field representation in the dark accurately. Specifically, frame-level holistic supervision, event-level granular supervision, and mixed-modality sharpening supervision. At the first level, Dark-EvGS leverages pseudo-bright frames, though it may be inaccurate in detail, to obtain a holistic view of knowledge through frame-level supervision. At the second level, filtered events serve as a supervisory signal to refine granular details that are ignored in previously used frames. Finally, at the third level, to mitigate motion blur effects in dark scenes, a combined event and frame supervision approach is applied using a mixed-modality sharpening strategy.


\noindent\textbf{Frame-based Holistic Supervision.}
Pure events supervision brings additional noise under low lights and does not lead to good reconstruction results (Fig. \ref{fig:visual comparison} (f)). Though the pseudo bright light frame, generated by CTCM, may contain blur and minor defects in details, as ground truth, it provides a holistic overview of how the object in the current frame looks and supervises the rendered frame so it follows that high-level holistic framework.

At a time window with range $t_1$ and $t_2$, given two low-light dark frames $I^{t_1}_{low}$ and $I^{t_2}_{low}$ captured at these two timestamps, and the ground truth accumulated events $E_{gt}$ between the time window, the Color Tone Consistency
Module, CTCM, generates two estimated bright frames $B^{t_1}$, and $B^{t_2}$, where $(B^{t_1}, B^{t_2}) = \text{CTCM}(I^{t_1}_{low}, I^{t_2}_{low}, E_{gt})$. We then use $N$ rendered frames from the GS model to compute $\mathcal{L}2$ loss with an estimated bright frame $B$, where $N$ is the number of viewpoints. Therefore, the frame-based holistic loss is defined as:
\begin{equation} 
\mathcal{L}_{hol}(I, B) = \frac{1}{N} \sum_{i=1}^{N} (B_i - I_i)^2.
\end{equation}

\noindent\textbf{Event-level Granular Supervision.}
Granular details are ignored by the frame camera when captured in the dark due to limited dynamic range. Furthermore, the estimated pseudo-bright frame from CTCM can only give a holistic view, ignoring the same region of sharp and accurate details. Event streams captured by the event camera play a critical role in supervising details and minors because they are high in temporal resolution, so details between frames can be captured, and they are wide in dynamic range, so unrevealed information can be seen. We use a y-noise filter \cite{feng2020event} $Y$ to form ground truth event $E_{gt} = Y(E_{raw})$ from raw captured events $E_{raw}$. To learn $E_{pred}$ with the formulated clean event signals $E_{gt}$, we adopt the classic event supervision loss from \cite{rudnev2023eventnerf}, where $N$ is the number of rendering viewpoints:
\begin{equation} 
\mathcal{L}_{event}(E_{pred}, E_{gt}) = \frac{1}{N} \sum_{i=1}^{N} (E^i_{pred} - E^i_{gt})^2.
\end{equation}
This helps to refine details that the frame-based holistic supervision failed to reconstruct. 

\begin{figure}[t]
	
	\centering
	
	\includegraphics[width=\linewidth,scale=1.0]{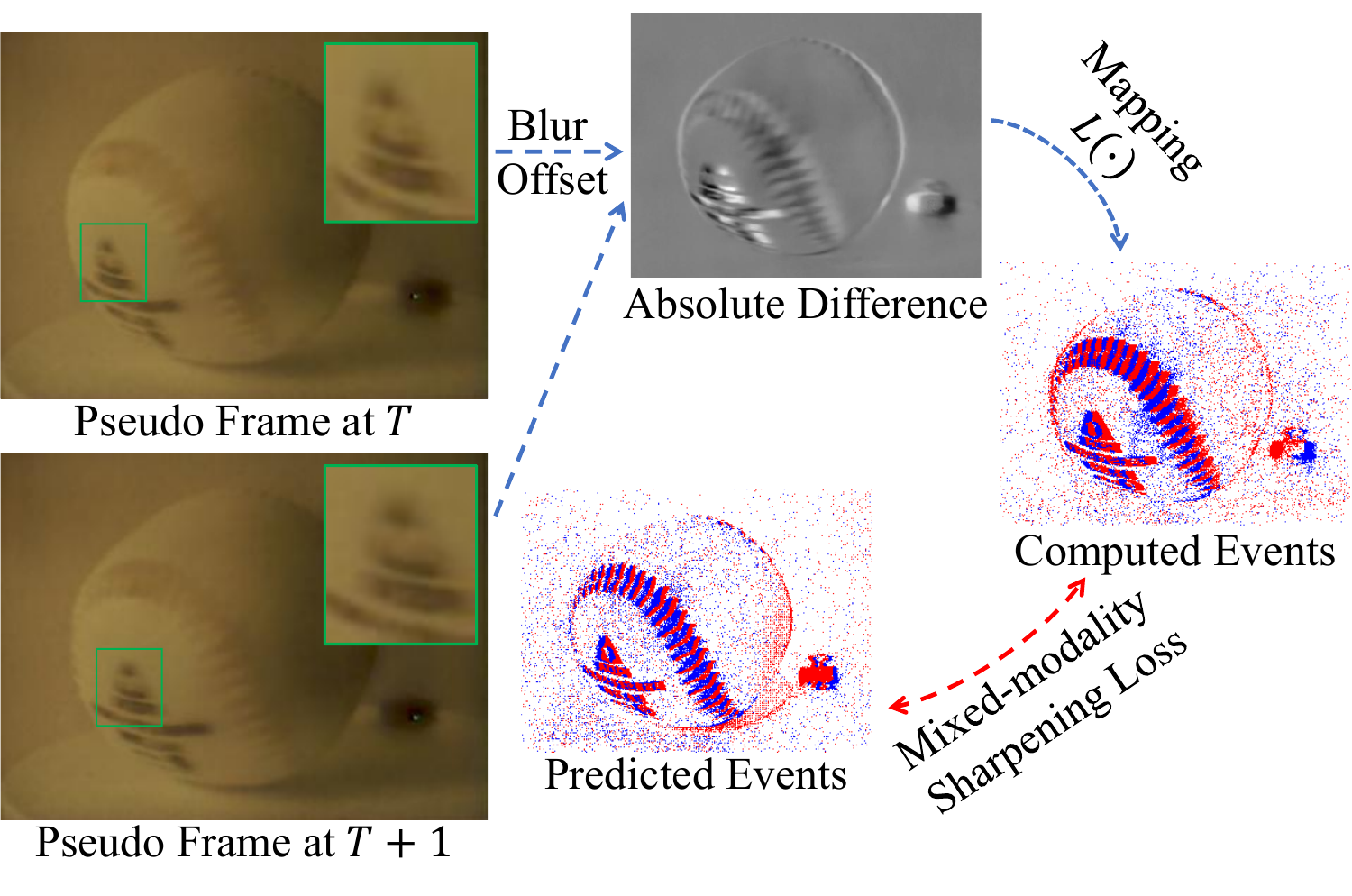}
	
	\caption{Illustration of our proposed mix-modality sharpening loss. The blurred regions (green boxes) can be offset, thus producing sharper rendering results.}
	
	\label{fig:mix loss}
	
\end{figure}

\noindent\textbf{Mixed-Modality Sharpening Supervision.}
Because the generated pseudo-bright frames from CTCM may contain motion blur, applying a strict $\mathcal{L}2$ loss leads to the rendered results being blurry as well. To correct motion blur in the rendering process, we propose a mixed-modality sharpening loss. The key idea is: though individual generated bright frames from CTCM contain motion blur, consecutive frames retain temporal consistency - motion blur and sharp areas of nearby frames belong to similar area and thus can be compensated by subtraction (Fig. \ref{fig:mix loss} green boxes). Mathematically speaking, the subtraction (Fig. \ref{fig:mix loss} absolute difference) between two generated bright frames is sharp and stands for the pixel shift in spatial. Thus, it can be a good guide for the spatial shift to rendered GS results. Therefore, we map the sharp subtraction results using the logarithmic function $L$, to formulate the computed event signal and then supervise the predicted events from the 3D GS network. This effectively eliminates the blurring effect and produces sharper results during rendering. Here, $I^{t_1}$ and $I^{t_2}$ denote two rendered intensity images from the 3D Gaussian Splatting model at timestamps $t_1$ and $t_2$, respectively, while $B^{t_1}$ and $B^{t_2}$ are the corresponding pseudo-bright frames generated by the CTCM. The operator $L(\cdot)$ represents the logarithmic intensity mapping used to compute log-intensity differences. The index $i$ enumerates frames, and $N$ denotes the total number of frames. The mixed-modality sharpening loss $\mathcal{L}_{mix}$ is defined as:

\begin{center}
\begin{equation} 
\begin{split}
\mathcal{L}_{mix}(I^{t_1}, I^{t_2},& B^{t_1}, B^{t_2}) = \\  \frac{1}{N} \sum_{i=1}^{N} ((L(I^{t_1}) - L(I^{t_2})) &- (L(B^{t_1}) - L(B^{t_2})))^2.
\end{split}
\end{equation}
\end{center}

\noindent\textbf{Final Loss.}
The final loss function is expressed as:
\begin{equation} 
\mathcal{L} = \mathcal{L}_{hol} + \lambda_1 \mathcal{L}_{event} + \lambda_2 \mathcal{L}_{mix}, 
\label{Eq: loss}
\end{equation} 
where $\lambda_1$ and $\lambda_2$ is set to $0.25$ across all experiments.

Noise filtering using the $y$-noise filter effectively removes interfering signals, especially in dark and low-light, while the triplet supervision keeps the training stable across varying conditions. This stability ensures consistent rendering performance, regardless of data source.

\section{Experiments}
\label{sec:results}
This section provides details of our data collection, experiment setup, evaluation results, and analysis.

\subsection{Data Collection}
\label{sec: data collection}

\begin{figure}[h]
	
	\centering
	
	\includegraphics[width=1\linewidth]{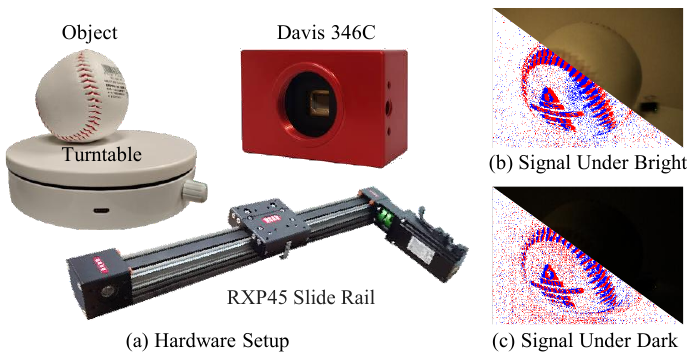}
	
\caption{Demonstration of hardware setup and dataset collection. 
Objects are captured either on a turntable or using a linear translation stage, with an event camera observing the scene (a). 
Signals are collected under both bright (b) and dark (c) illumination conditions. 
In low-light environments, frame-based images suffer from severe information loss and motion blur, while event signals become noisy.}

	\label{fig:setup}
	
\end{figure}

In this subsection, we introduce the real-world data collection process in detail. 
Fig. \ref{fig:setup} (a) demonstrates the hardware setup. Our real-world dataset consists of two complementary parts: (i) single-object turntable scenes and (ii) forward-facing scenes with complex objects. 
The former enables controlled analysis under low-light conditions, while the latter is introduced to evaluate the generalization ability of the proposed method beyond turntable-based settings.

\paragraph{Single-object turntable scenes}
To collect the single-object dataset, we place a static object on a constant, unknown-speed turntable. 
A Davis 346 Color event camera is held steady and points toward the object throughout the capture process. 
We use a TES 1339R light meter to measure the illumination level of the environment in lux. 
Six single-object scenes are collected, and for each scene, we set two lighting conditions to form paired data: bright (around 300 lux) and dark (less than 40 lux). 
All scenes are fully covered by curtains to eliminate natural light interference, and a controllable artificial light source is placed directly above the object.

Under bright lighting, we use the highest illumination level and capture the corresponding frames, event streams, and camera information during a full 360-degree rotation. 
Subsequently, we repeat the same mechanically controlled rotation trajectory under low-light conditions, leaving only limited illumination on the object, and capture the same set of data. 
Among the six scenes, three are categorized as moderate lighting (between 20 and 40 lux, including ``Baseball'', ``House'', and ``Lion''), and three are categorized as challenging lighting (less than 20 lux). 
For each scene, paired dark–bright frames, event streams, and camera poses are prepared. 
Bright-light frames are used as ground-truth supervision for evaluation, while only dark frames and/or dark events are available during training and supervision.

Fig. \ref{fig:setup} (b) and (c) show example data captured under bright and dark lighting conditions, respectively. 
As illustrated, frames captured under normal lighting are sharp and informative, while frames captured under low-light conditions suffer from severe information loss and motion blur, making them difficult to interpret for both humans and machines. 
At the same time, event streams captured in dark environments exhibit significantly higher noise levels.

\begin{figure*}[h]
	
	\centering
	
	\includegraphics[width=\linewidth,scale=1.0]{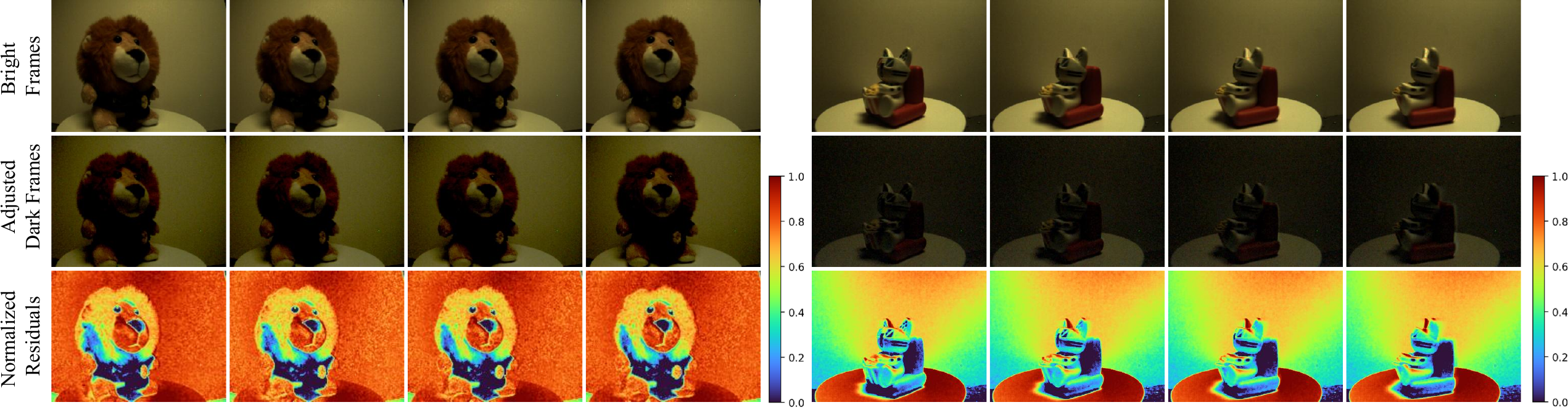}
	
	\caption{
Paired bright and dark frame samples from the proposed real-world dataset.
The first row shows bright frames captured under normal illumination.
The second row shows the corresponding dark frames after brightness and contrast adjustment
(\,+40\% brightness, \,-40\% contrast\,), which is applied only for visualization and residual analysis to better reveal structural correspondence.
The third row visualizes the residual heatmaps computed between each bright frame and its adjusted dark counterpart.
As shown, the residuals are dominated by global intensity differences, while object contours and structural edges remain well aligned,
indicating that the paired bright–dark frames are temporally synchronized and pixel-wise aligned, differing primarily in illumination conditions
rather than spatial misalignment.}

	\label{fig:pair vis}
	
\end{figure*}

A visual illustration of paired bright and dark frames is provided in Fig.~\ref{fig:pair vis}, where each bright–dark pair corresponds to the same camera viewpoint along the rotation trajectory. 
Since both sequences are captured using the same camera with fixed intrinsics and an identical mechanically controlled motion, the paired frames are temporally synchronized and pixel-wise aligned, differing only in illumination conditions.

\paragraph{Forward-facing scenes with complex objects}
While turntable-based capture is commonly adopted in radiance field reconstruction for controlled benchmarking, it does not fully reflect forward-facing camera motion encountered in practical scenarios. To address this limitation, we additionally collect forward-facing scenes with complex objects using the RXP45 Slide Rail linear translation stage (shown in Fig. \ref{fig:setup} (a)). In this setting, the event camera undergoes a translational motion along a linear trajectory while observing cluttered scenes composed of multiple objects and background structures. This setup simulates forward-facing viewpoints and unbounded scene configurations. The translation stage is controlled by an Arduino-based pulse generator. The stepper driver is configured with $6400$ pulses per revolution, and the stage is driven at a constant speed of $20$~RPM or $2.67 \mathrm{cm/s}$. During acquisition, the camera is placed on a fixed location of the translation stage, moving along with it, and illumination remains the same, and we record event streams and frames at a constant sensor configuration. These forward-facing scenes are captured under low-light conditions following the same acquisition protocol, and the corresponding bright-light frames are used solely for evaluation. By incorporating this additional data, we aim to demonstrate that the proposed framework is not limited to turntable-based scenarios but can also generalize to more realistic camera trajectories and complex scene layouts.

\subsection{Implementation Details}
Our implementation builds upon the 3D GS framework \cite{kerbl20233d}, leveraging its primary structure and functionalities. As 3D GS requires a point cloud input, we randomly initialize $10^3$ points to create the starting point cloud, as the structure-from-motion initialization \cite{schonberger2016structure} in the original setup is not directly applicable to event data. We use the pre-trained weights from EvLowLight on its existing modules with randomly initialized CTMB weights as a start. They are then tuned with five hundred unrelated real-world dark-bright data pairs and corresponding events \footnote{These are extra-captured natural scene data using Davis 346C.} for 5 epochs to provide robust prior knowledge on pseudo frames. All experiments were conducted on an NVIDIA RTX 3090 GPU. We used specific hyperparameters to ensure optimal performance. Training was conducted for a total of 30,000 iterations for all scenes. For position optimization, the learning rate was scheduled to decay from $1.6 \times 10^{-4}$ to $1.6 \times 10^{-6}$. The learning rates for optimizing features, opacity, scaling, and rotation were set to $2.5 \times 10^{-3}$, $5 \times 10^{-2}$, $5 \times 10^{-3}$, and $1 \times 10^{-3}$, respectively.

\begin{table*}[h]
\centering
\small
\caption{Quantitative comparison on the real dataset. Metrics include PSNR, SSIM, and LPIPS across various scenes. 
$\uparrow$ means higher is better and $\downarrow$ means lower is better. 
The first column (\textbf{Input}) indicates the input modality: \textbf{F} = frame only, \textbf{E} = event only, \textbf{E+F} = event + frame. 
For methods that are not designed for radiance field reconstruction (e.g., E2VID and EvLowLight), their enhanced outputs are used as input to a subsequent 3D Gaussian Splatting stage for 3D reconstruction and evaluation. Results marked with * indicate models fine-tuned on the same real-world dark–bright data pairs used for CTCM training. 
Best results are in \textbf{bold} and second best are \underline{underlined}.}
\label{compare PSNR SSIM LPIPS}
\begin{tabular}{cc|ccc|ccc|ccc}
\toprule
\multicolumn{11}{c}{\textbf{Moderate Lighting Scene (20 - 40 lux)}}\\
\toprule
\multirow{2}{*}{\textbf{Input}} & \multirow{2}{*}{\textbf{Method}} & \multicolumn{3}{c|}{Baseball (28 lux)} & \multicolumn{3}{c|}{House (38 lux)} & \multicolumn{3}{c}{Lion (32 lux)} \\
 &  & PSNR$\uparrow$ & SSIM$\uparrow$ & LPIPS$\downarrow$ & PSNR$\uparrow$ & SSIM$\uparrow$ & LPIPS$\downarrow$ & PSNR$\uparrow$ & SSIM$\uparrow$ & LPIPS$\downarrow$ \\
\midrule
\multirow{1}{*}{F}   & 3D GS \cite{kerbl20233d}            & 9.91 & 0.15 & 0.52 & 8.79 & 0.21 & 0.55 & 9.32  & 0.23 & 0.58 \\
\cline{1-1}
\multirow{4}{*}{E}   & E2VID \cite{rebecq2019events}       & 9.78 & 0.37 & 0.65 & 9.32 & 0.49 & 0.69 & 9.54  & 0.46 & 0.72 \\
                    & E2VID* & 11.24 & 0.41 & 0.64 & 10.62 & 0.50 & 0.67 & 9.93  & 0.47 & 0.71 \\
                     & EventNeRF \cite{rudnev2023eventnerf}& 9.05 & 0.59 & 0.58 & 10.06& 0.62 & 0.67 & 10.43 & 0.60 & 0.68 \\
                     & EvGS \cite{wu2024EV-GS}             & 15.82& 0.72 & 0.51 & 13.27& 0.67 & 0.57 & 14.95 & 0.68 & 0.60 \\
\cline{1-1}
\multirow{5}{*}{E+F} & E2GS \cite{deguchi2024e2gs}         & 14.90& 0.70 & 0.55 & 14.31& 0.68 & 0.55 & 14.19 & 0.70 & 0.62 \\
                     & SweepEvGS \cite{wu2025sweepevgs}    & 20.25& 0.79 & 0.40 & 19.29& 0.72 & 0.41 & \underline{20.37} & \underline{0.75} & \underline{0.45} \\
                     & EvLowLight \cite{liang2023coherent} & 20.34 & \underline{0.83} & 0.38 & \underline{19.31} & \underline{0.75} & \underline{0.45} & 18.86 & 0.74 & 0.49 \\
                     & EvLowLight* & \underline{21.02} & 0.81 & \underline{0.37} & 19.17 & 0.70 & 0.48 & 18.94 & 0.74 & 0.47 \\
                     & Ours                                & \textbf{26.58} & \textbf{0.87} & \textbf{0.31} & \textbf{23.26} & \textbf{0.81} & \textbf{0.36} & \textbf{31.22} & \textbf{0.85} & \textbf{0.30} \\
\toprule
\multicolumn{11}{c}{\textbf{Challenging Lighting Scene (\textless 20 lux)}}\\
\toprule
\multirow{2}{*}{\textbf{Input}} & \multirow{2}{*}{\textbf{Method}} & \multicolumn{3}{c|}{Panda (17 lux)} & \multicolumn{3}{c|}{Badminton (14 lux)} & \multicolumn{3}{c}{Cat (16 lux)} \\
 &  & PSNR$\uparrow$ & SSIM$\uparrow$ & LPIPS$\downarrow$ & PSNR$\uparrow$ & SSIM$\uparrow$ & LPIPS$\downarrow$ & PSNR$\uparrow$ & SSIM$\uparrow$ & LPIPS$\downarrow$ \\
\midrule
\multirow{1}{*}{F}   & 3D GS \cite{kerbl20233d}            & 8.73 & 0.08 & 0.69 & 14.69 & 0.58 & 0.35 & 10.61 & 0.09 & 0.46 \\
\cline{1-1}
\multirow{4}{*}{E}   & E2VID \cite{rebecq2019events}       & 9.38 & 0.38 & 0.70 & 8.26  & 0.15 & 0.65 & 8.92  & 0.35 & 0.64 \\
                     & E2VID* & 10.14 & 0.40 & 0.67 & 9.12  & 0.31 & 0.65 & 9.15  & 0.39 & 0.60 \\
                     & EventNeRF \cite{rudnev2023eventnerf}& 9.23 & 0.56 & 0.69 & 5.91  & 0.14 & 0.48 & 5.95  & 0.33 & 0.46 \\
                     & EvGS \cite{wu2024EV-GS}             & 12.37& 0.63 & 0.60 & 10.89 & 0.17 & 0.35 & 10.58 & 0.21 & 0.37 \\
\cline{1-1}
\multirow{5}{*}{E+F} & E2GS \cite{deguchi2024e2gs}         & 14.20& 0.69 & 0.50 & 13.72 & 0.68 & 0.60 & 11.73 & 0.27 & 0.39 \\
                     & SweepEvGS \cite{wu2025sweepevgs}    & \underline{17.22} & 0.73 & 0.47 & 18.84 & \underline{0.33} & 0.28 & 18.56 & 0.72 & 0.45 \\
                     & EvLowLight \cite{liang2023coherent} & 16.28 & \underline{0.76} & \underline{0.46} & 18.87 & 0.31 & \underline{0.20} & \underline{19.67} & \underline{0.78} & \underline{0.38} \\
                     & EvLowLight* & 16.51 & 0.75 & 0.48 & \underline{19.20} & 0.32 & 0.22 & 19.23 & 0.74 & 0.40 \\
                     & Ours                                & \textbf{20.84} & \textbf{0.80} & \textbf{0.36} & \textbf{23.70} & \textbf{0.42} & \textbf{0.19} & \textbf{21.64} & \textbf{0.83} & \textbf{0.33} \\
\toprule
\end{tabular}
\end{table*}

\begin{table}[t]
\centering
\small
\setlength{\tabcolsep}{0.35mm}
\caption{Quantitative comparison on forward-facing scenes with complex objects.
Metrics include PSNR, SSIM, and LPIPS. 
$\uparrow$ indicates higher is better and $\downarrow$ indicates lower is better. 
The first column (\textbf{Input}) indicates the input modality: \textbf{F} = frame only, 
\textbf{E} = event only, and \textbf{E+F} = event + frame. 
For methods not originally designed for radiance field reconstruction (e.g., E2VID and EvLowLight), 
their enhanced outputs are fed into a subsequent 3D Gaussian Splatting stage for reconstruction. 
Results marked with * indicate models fine-tuned on the same real-world data pairs used for CTCM training. 
Best results are shown in \textbf{bold} and second-best results are \underline{underlined}.}
\label{compare_forward_facing}

\begin{tabular}{c c|ccc|ccc}
\toprule
\multirow{2}{*}{\textbf{Input}} & \multirow{2}{*}{\textbf{Method}} 
& \multicolumn{3}{c|}{Toys (30 lux)} 
& \multicolumn{3}{c}{Chess (15 lux)} \\
 &  & PSNR$\uparrow$ & SSIM$\uparrow$ & LPIPS$\downarrow$
 & PSNR$\uparrow$ & SSIM$\uparrow$ & LPIPS$\downarrow$ \\
\midrule
\multirow{1}{*}{F}
& 3D GS
& 10.12 & 0.22 & 0.58
& 9.84  & 0.19 & 0.60 \\
\cline{1-1}

\multirow{4}{*}{E}
& E2VID
& 10.45 & 0.39 & 0.66
& 10.02 & 0.36 & 0.68 \\
& E2VID*
& 14.36 & 0.63 & 0.53
& 13.88 & 0.55 & 0.55 \\
& EventNeRF
& 10.21 & 0.55 & 0.60
& 9.76  & 0.52 & 0.62 \\
& EvGS
& 11.92 & 0.48 & 0.59
& 11.37 & 0.45 & 0.57 \\
\cline{1-1}

\multirow{5}{*}{E+F}
& E2GS
& 15.31 & 0.71 & 0.46
& 14.85 & 0.69 & 0.48 \\
& SweepEvGS
& 18.74 & 0.77 & 0.41
& 18.29 & 0.75 & 0.43 \\
& EvLowLight
& 18.96 & \underline{0.79} & \underline{0.37}
& 18.12 & \underline{0.77} & \underline{0.41} \\
& EvLowLight*
& \underline{19.42} & 0.78 & 0.40
& \underline{18.65} & 0.74 & 0.43 \\
& Ours
& \textbf{22.85} & \textbf{0.84} & \textbf{0.30}
& \textbf{21.97} & \textbf{0.82} & \textbf{0.32} \\
\bottomrule
\end{tabular}
\end{table}

\begin{table}[h]
\centering
\small
\setlength{\tabcolsep}{1mm}
\caption{Comparison of training time (Train), rendering performance (FPS), and GPU memory usage (Mem) across different methods on Moderate Lighting Scene and Challenging Lighting Scene. The metrics include training time (minutes and hours), frames per second (FPS) for real-time rendering, and GPU memory usage (GB) during training. $\uparrow$ indicates that higher values are better, while $\downarrow$ indicates that lower values are better. The metrics in bold are ranked first, and the \underline{underlined} metrics are ranked second.}
\begin{tabular}{r|rrr|rrr} 
\toprule
\multirow{2}{*}{Method} & \multicolumn{3}{c|}{Moderate Lighting} & \multicolumn{3}{c}{Challenging Lighting}\\
& Train$\downarrow$ & FPS$\uparrow$ & Mem$\downarrow$ & Train$\downarrow$ & FPS$\uparrow$ & Mem$\downarrow$ \\ \toprule
3D GS  & \textbf{3min} & \underline{40} & \textbf{1GB} & \textbf{3min} & \underline{32} & \textbf{1.1GB} \\
E2VID  & 5min & 35 & 1.8GB & 5min & 30 & 1.8GB \\
EventNeRF  & 19h & 0.5 & 12GB & 26h & 0.3 & 12GB \\
EvGS  & 4.5min & 33 & 1.5GB & 4.5min & 31 & 2.1GB \\ 
E2GS  & 30min & 30 & 5GB & 32min & 28 & 5GB \\  
SweepEvGS  & 8min & 38 & 2GB & 8min & 40 & 2GB \\  
EvLowLight  & 11h & 3.5 & 2GB & 10h & 3.5 & 3GB \\
Ours & \underline{4min} & \textbf{41} & \underline{1.2GB} & \underline{4min} & \textbf{35} & \underline{1.7GB} \\ \toprule
\end{tabular}
\label{compare efficiency}
\end{table}

\subsection{Quantitative Evaluation}
We compare the radiance field reconstruction quality and efficiency of various methods, including frame-based method (F): 3D GS \cite{kerbl20233d}; pure-event based methods (E): E2VID \cite{rebecq2019events}, EventNeRF \cite{rudnev2023eventnerf}, and Ev-GS \cite{wu2024EV-GS}; and hybrid methods (E+F): EvLowLight \cite{liang2023coherent}, Sweep-EvGS \cite{wu2025sweepevgs}, and E2GS \cite{deguchi2024e2gs}, on our real-world dataset. For approaches that do not directly perform radiance field reconstruction (e.g., E2VID and EvLowLight), their enhanced frame outputs are further processed by a 3D Gaussian Splatting pipeline to obtain the final 3D reconstruction results used for evaluation. To further ensure a fair comparison, we additionally report results of fine-tuned baseline methods (denoted by *), where E2VID and EvLowLight are fine-tuned on the same type of auxiliary real-world dark–bright data pairs as CTCM training, with no overlap with the evaluation scenes.
Following the standard protocol, we adopt an $8:2$ train-test split to ensure reliable evaluation across both moderate and challenging lighting scenes. The evaluation considers the following two key aspects:  

(i) \textit{Reconstruction Quality:} We use PSNR, SSIM, and LPIPS as metrics to quantify the fidelity of reconstructed images compared to the bright-light ground truth. Higher PSNR and SSIM values correspond to better reconstruction in terms of brightness and structural similarity, while lower LPIPS values indicate better perceptual similarity. As reported in Table~\ref{compare PSNR SSIM LPIPS} and Table \ref{compare_forward_facing}, our method consistently achieves the best performance across all scenes. For example, in the \textit{Lion} scene under moderate conditions, our approach improves PSNR to $31.22$ and SSIM to $0.85$, outperforming the second-best EvLowLight \cite{liang2023coherent} by a clear margin, while reducing LPIPS to $0.30$. Similar advantages are observed in challenging lighting scenes, such as the \textit{Cat} sequence, where our method achieves the highest perceptual similarity and structural fidelity.

(ii) \textit{Efficiency:} To evaluate practical usability, we measure training time, rendering FPS, and GPU memory consumption. As shown in Table~\ref{compare efficiency}, our method strikes a favorable balance between efficiency and accuracy. Specifically, Dark-EvGS requires only $4$ min of training, significantly faster than hybrid methods such as E2GS \cite{deguchi2024e2gs} ($30$ min) and Sweep-EvGS \cite{wu2025sweepevgs} ($8$ min), while achieving the highest rendering FPS ($41$ and $35$ FPS in moderate and challenging lighting scenes, respectively). In terms of GPU memory usage, our framework also remains lightweight ($1.2$ to $1.7$ GB), close to the most efficient 3D GS \cite{kerbl20233d} baseline, and much lower than NeRF-based approaches such as EventNeRF \cite{rudnev2023eventnerf} ($12$ GB).  

Overall, these results highlight that Dark-EvGS is not only more accurate in reconstructing radiance fields under dark environments but also practical for real-time deployment, owing to its fast training, high rendering speed, and modest memory requirements.

\subsection{Ablation Studies}
\label{supp: Ablation}
We conduct ablation studies on the effect of rendering quality from the following three aspects: the supervision module and the event utilization module; we then provide the ablation study on the effect of efficiency from each component.

\subsubsection{Ablation Study on Supervision Module}
\begin{table*}[htbp!]
\centering
\caption{Detailed Ablation Study on Dark-EvGs. \checkmark denotes applied components. The results of PSNR and SSIM on four samples, adding each component individually and applying the components sequentially, are reported. \textbf{Bold} denotes the highest score.}
\setlength{\tabcolsep}{1.5mm}{
\renewcommand\arraystretch{1.3} {
\begin{tabular}{ccc|cc|cc|cc|cc|cc|cc}
\toprule[1.5px]
\multicolumn{1}{c}{\multirow{2}{*}{$\mathcal{L}_{hol}$}} & \multicolumn{1}{c}{\multirow{2}{*}{$\mathcal{L}_{event}$}} & \multicolumn{1}{c|}{\multirow{2}{*}{$\mathcal{L}_{mix}$}} & \multicolumn{2}{c|}{Baseball} & \multicolumn{2}{c|}{House} & \multicolumn{2}{c|}{Lion} & \multicolumn{2}{c|}{Panda} & \multicolumn{2}{c|}{Badminton} & \multicolumn{2}{c}{Cat} \\
& & & PSNR & SSIM & PSNR & SSIM & PSNR & SSIM & PSNR & SSIM & PSNR & SSIM & PSNR & SSIM\\ \hline \hline
\checkmark &            &            & 20.37 & 0.83 & 18.57 & 0.74 & 25.36 & 0.80 & 16.31 & 0.76 & 18.21 & 0.28 & 15.44 & 0.71 \\
           & \checkmark &            & 16.67 & 0.73 & 14.96 & 0.69 & 20.73 & 0.77 & 13.19 & 0.64 & 15.12 & 0.19 & 13.11 & 0.62 \\
           &            & \checkmark & 14.61 & 0.70 & 13.21 & 0.67 & 19.11 & 0.75 & 11.89 & 0.51 & 13.12 & 0.15 & 11.61 & 0.49 \\
\checkmark & \checkmark &            & 24.93 & 0.85 & 22.18 & 0.79 & 29.92 & 0.83 & 19.23 & 0.78 & 22.03 & 0.40 & 19.51 & 0.79 \\

\checkmark &            & \checkmark & 22.61 & 0.84 & 20.27 & 0.76 & 27.84 & 0.81 & 18.37 & 0.77 & 20.17 & 0.36 & 17.71 & 0.76 \\
           & \checkmark & \checkmark & 18.23 & 0.78 & 15.72 & 0.71 & 22.45 & 0.78 & 14.41 & 0.67 & 15.73 & 0.21 & 14.23 & 0.66 \\
\checkmark & \checkmark & \checkmark & \bf 26.58 & \bf 0.87 & \bf 23.26 & \bf 0.81 & \bf 31.22 & \bf 0.85 & \bf 20.84 & \bf 0.80 & \bf 23.70 & \bf 0.42 & \bf 21.64 & \bf 0.83 \\
\bottomrule[1.5px]
\end{tabular}
}}
\label{Detailed Ablation}
\end{table*}

To understand the individual contributions of each component in our proposed framework, Dark-EvGS, we conducted a detailed ablation study. Table \ref{Detailed Ablation} presents the results of the PSNR and SSIM metrics on four test samples, with various combinations of loss components applied: $\mathcal{L}_{hol}$, $\mathcal{L}_{event}$, and $\mathcal{L}_{mix}$. The study highlights the unique impact of each component, demonstrating their complementary roles in improving rendering quality.  

The holistic supervision loss, $\mathcal{L}_{hol}$, significantly improves global coherence, as shown by the notable increase in both PSNR and SSIM when applied alone. The event-based supervision loss, $\mathcal{L}_{event}$, contributes to capturing dynamic details, while the mix supervision loss, $\mathcal{L}_{mix}$, refines local textures. When combined, these components work synergistically, with $\mathcal{L}_{hol} + \mathcal{L}_{event}$ balancing global and dynamic details and the addition of $\mathcal{L}_{mix}$ enhancing fine-grained reconstruction. 

The full framework, combining all three components, achieves the best results across all samples, with the highest PSNR and SSIM values, such as $31.22$ and $0.85$ for the ``Lion'' scene. These results confirm that each component plays a vital role and that their combination maximizes rendering quality for low-light radiance field reconstruction.  

\subsubsection{Ablation Study on Utilization Module}
\begin{table*}[h]
\centering
\caption{Ablation Study on Event Noise Filtering with Different Temporal Window $\Delta T$. 
The results of PSNR and SSIM on six scenes are reported. \textbf{Bold} denotes the highest score.}
\setlength{\tabcolsep}{1.5mm}{
\renewcommand\arraystretch{1.2} {
\begin{tabular}{c|cc|cc|cc|cc|cc|cc}
\toprule[1.5px]
\multicolumn{1}{c|}{\multirow{2}{*}{Method}} & 
\multicolumn{2}{c|}{Baseball} & 
\multicolumn{2}{c|}{House} & 
\multicolumn{2}{c|}{Lion} & 
\multicolumn{2}{c|}{Panda} & 
\multicolumn{2}{c|}{Badminton} & 
\multicolumn{2}{c}{Cat} \\
 & PSNR & SSIM & PSNR & SSIM & PSNR & SSIM & PSNR & SSIM & PSNR & SSIM & PSNR & SSIM\\ 
\hline \hline

w/o Noise Filter 
& 24.23 & 0.85 
& 21.01 & 0.78 
& 29.78 & 0.83 
& 19.45 & 0.78 
& 20.22 & 0.37 
& 20.27 & 0.80 \\

with Noise Filter ($\Delta T=1000\,\mu$s) 
& 25.61 & 0.86 
& 22.35 & 0.80 
& 30.54 & 0.84 
& 20.12 & 0.79 
& 22.01 & 0.40 
& 21.08 & 0.82 \\

with Noise Filter ($\Delta T=2000\,\mu$s) 
& \bf 26.58 & \bf 0.87 
& \bf 23.26 & \bf 0.81 
& \bf 31.22 & \bf 0.85 
& \bf 20.84 & \bf 0.80 
& \bf 23.70 & \bf 0.42 
& \bf 21.64 & \bf 0.83 \\

with Noise Filter ($\Delta T=3000\,\mu$s) 
& 25.94 & 0.86 
& 22.71 & 0.80 
& 30.88 & 0.84 
& 20.31 & 0.79 
& 22.64 & 0.41 
& 21.21 & 0.82 \\

with Noise Filter ($\Delta T=4000\,\mu$s) 
& 23.81 & 0.84 
& 20.36 & 0.77 
& 28.92 & 0.82 
& 18.97 & 0.77 
& 19.45 & 0.35 
& 19.88 & 0.79 \\

\bottomrule[1.5px]
\end{tabular}
}}
\label{Noise Filter Ablation}
\end{table*}


\begin{table}[h]
\centering
\small
\setlength{\tabcolsep}{1mm}
\caption{Efficiency ablation on Dark-EvGS. \checkmark\ denotes applied components.}
\label{Efficiency Ablation}
\begin{tabular}{ccc|cc|cc}
\toprule
\multicolumn{3}{c|}{Loss config} & \multicolumn{2}{c|}{Moderate Lighting} & \multicolumn{2}{c}{Challenging Lighting} \\
$\mathcal{L}_{\text{hol}}$ & $\mathcal{L}_{\text{event}}$ & $\mathcal{L}_{\text{mix}}$ & Time (min)$\downarrow$ & FPS$\uparrow$ & Time (min)$\downarrow$ & FPS$\uparrow$ \\ \toprule
\checkmark &            &            & 3.2 & 35 & 3.2 & 35 \\
           & \checkmark &            & 3.5 & 35 & 3.5 & 35 \\
           &            & \checkmark & 3.0 & 40 & 3.0 & 43 \\
\checkmark & \checkmark &            & 3.8 & 35 & 3.7 & 30 \\
\checkmark &            & \checkmark & 3.5 & 38 & 3.4 & 38 \\
           & \checkmark & \checkmark & 3.7 & 35 & 3.6 & 40 \\
\checkmark & \checkmark & \checkmark & 4.0 & 41 & 4.0 & 35 \\ \toprule
\end{tabular}
\end{table}

To evaluate the effectiveness of the proposed noise filter in the event utilization module, we conducted an ablation study comparing the performance of our framework with and without the noise filter. In this work, we adopt the y-noise filter from \cite{feng2020event}, which is specifically designed to suppress background activity noise while preserving meaningful event structures. The core idea is that noise events tend to appear sparsely and randomly in both space and time, whereas signal events form locally dense clusters corresponding to real intensity changes. Concretely, raw events are first grouped within a short temporal window $\Delta T$. For each event, a local spatial neighborhood of size $L \times L$ is considered to construct an event density map. Events with a local density below a predefined threshold are considered noise and are removed. This process effectively filters out isolated events while retaining temporally and spatially coherent event patterns generated by actual scene motion or intensity variation. In our ablation study, we further investigate the impact of the temporal window size $\Delta T$ on reconstruction performance, as it directly controls the trade-off between noise suppression and detail preservation in event density estimation.

Table~\ref{Noise Filter Ablation} reports the PSNR and SSIM results on six test scenes under different noise filtering settings. The results clearly demonstrate the necessity of incorporating event noise filtering. Without the noise filter, the framework struggles to suppress background noise, leading to degraded reconstruction quality across all scenes. For example, on the ``Baseball'' scene, the model achieves a PSNR of $24.23$ and an SSIM of $0.85$ without filtering, which improves significantly to $26.58$ and $0.87$, respectively, when an appropriate noise filter is applied. Moreover, the temporal window size $\Delta T$ plays a critical role in filtering effectiveness. When $\Delta T$ is too small (e.g., $1000\,\mu$s), noise is not sufficiently suppressed, resulting in limited performance gains. Conversely, an overly large temporal window (e.g., $4000\,\mu$s) leads to over-smoothing and loss of informative event structures, even degrading performance below the no-filter baseline. Among all settings, $\Delta T=2000\,\mu$s consistently achieves the best results across all scenes, indicating an optimal balance between noise suppression and detail preservation.

Overall, these results highlight the critical role of event noise filtering in stabilizing event supervision and refining fine-grained details, thereby improving global coherence and texture fidelity in the reconstructed radiance fields under extreme low-light conditions.

\subsubsection{Ablation Study on Efficiency}
To evaluate the efficiency of Dark-EvGS, we conduct an ablation study focusing on the computational cost introduced by each component. Table \ref{Efficiency Ablation} presents the training time (in minutes) and frames per second (FPS) rendered during inference for various configurations of the loss terms. 

Our analysis highlights that each component is computationally lightweight and introduces minimal overhead. For example, applying $\mathcal{L}_{hol}$ alone achieves training times of approximately $3.2$ minutes and maintains a steady FPS across all samples. Adding $\mathcal{L}_{event}$ or $\mathcal{L}_{mix}$ results in a slight increase in training time but ensures the FPS remains consistent or only marginally reduced. When all components are applied jointly, the total training time remains well within practical limits (approximately 4 minutes), while the FPS remains competitive, reaching up to $41$ FPS for moderate lighting scenes. These results demonstrate that the design of Dark-EvGS achieves a favorable trade-off between computational cost and efficiency. Each loss component contributes unique benefits without significantly increasing the computational burden, making the model both effective and resource-efficient.  

\subsection{Qualitative Evaluation}
\subsubsection{Qualitative Visualization}
\begin{figure*}[h]
	
	\centering
	
	\includegraphics[width=\linewidth,scale=1.0]{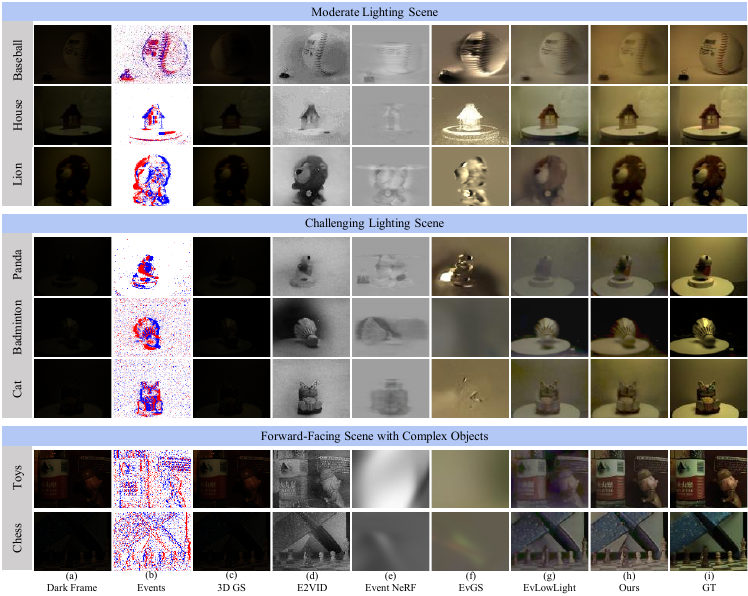}
	
	\caption{A visual comparison of different approaches on different scenes captured: 
(a) dark frames captured by a frame-based camera under low lights; 
(b) filtered events captured under low lights by an event camera; 
(c) results from 3D GS \cite{kerbl20233d} using dark frames; 
(d) rendering results from 3D GS using reconstructed frames from E2VID \cite{rebecq2019events}; 
(e) results from EventNeRF \cite{rudnev2023eventnerf}; 
(f) results from Ev-GS \cite{wu2024EV-GS}; 
(g) results from Ev-LowLight; 
(h) results from ours; 
(i) ground truth frames captured under bright lights. 
The top two rows correspond to object-centric scenes under moderate and challenging low-light conditions, respectively, while the bottom row shows forward-facing scenes with complex objects captured using a translational camera trajectory. 
Our approach renders brighter and sharper results purely using signals captured under dark environments compared to existing approaches across all scenes.}

	\label{fig:visual comparison}

\end{figure*}

\begin{figure*}[htbp]
    \centering
    \includegraphics[width=\linewidth,scale=1.0]{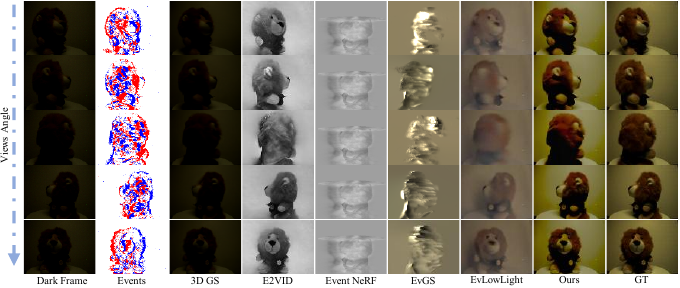}
    \caption{Multi-view comparison of rendered results from different methods on a low-light scene. For each viewpoint, Dark-EvGS consistently produces sharper, brighter, and more structurally accurate frames compared to baseline methods, demonstrating its effectiveness in reconstructing radiance fields under low-light conditions across diverse camera perspectives.}
    \label{fig:compare multi angle}
\end{figure*}

\begin{figure*}[htbp]
    \centering
    \includegraphics[width=\linewidth,scale=1.0]{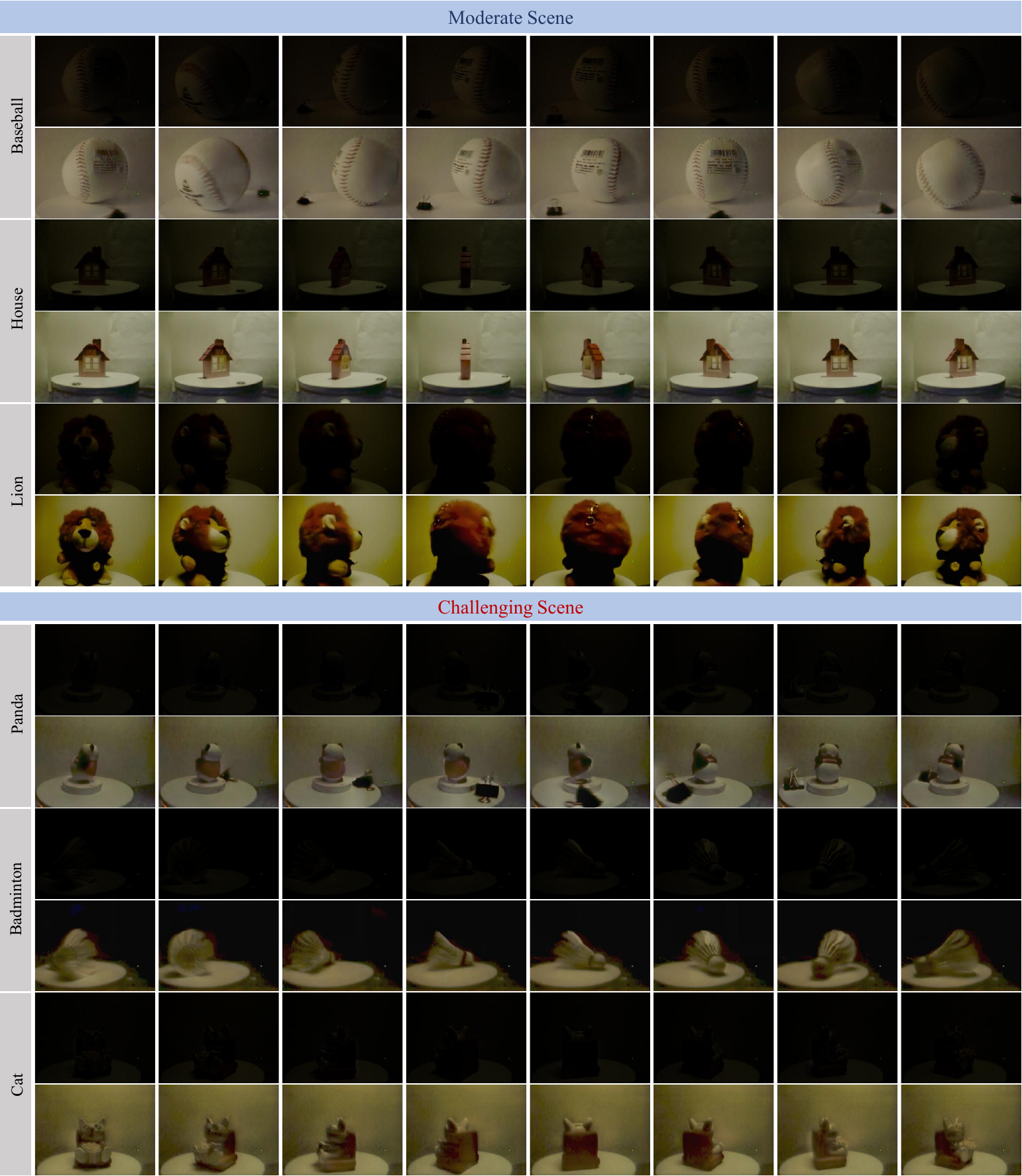}
    \caption{Visual comparison showcasing the performance of our method across all six real-world scenes (``Baseball", ``House", ``Lion" as moderate lighting scenes, and ``Panda", ``Badminton", ``Cat" as challenging lighting scenes). Each scene is viewed from multiple perspectives to comprehensively evaluate our method's ability to restore details, structure, and brightness under low-light conditions. In each example, the first row presents the input images captured in dark environments. In contrast, the second row shows the high-quality reconstructions generated by our approach, highlighting significant enhancements in visual clarity and realism.}
    \label{fig:More Vis}
\end{figure*}

We provide a comprehensive qualitative evaluation here. As shown in Fig.~\ref{fig:visual comparison}, frames captured in low light lose detail, while event data preserves fine structures regardless of illumination. Leveraging this advantage, our method produces clearer, more detailed renderings that better match the bright reference views. 
We also provide more visualization of different views for all samples in Fig. \ref{fig:More Vis}, and conduct a multi-view visualizations comparison against other methods in Fig. \ref{fig:compare multi angle}. For a given scene, we render outputs from multiple viewpoints along the camera trajectory using each competing method. The results consistently show that Dark-EvGS produces sharper, brighter, and more geometrically consistent frames across all angles. Compared to baseline methods, our reconstructions maintain clearer structural details and better visual fidelity under varying perspectives, highlighting the robustness and superiority of our approach in low-light radiance field reconstruction.

\begin{figure*}[htbp]
    \centering
    \includegraphics[width=\linewidth,scale=1.0]{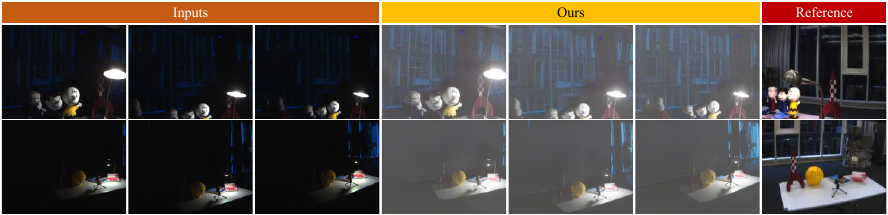}
    \caption{Qualitative evaluation of our approach on the open benchmark \cite{hidalgo2022event}. The orange rows represent the dark input frames, while the yellow rows show our rendered results. The red column contains bright reference frames from the same scene, which are unpaired and used only for qualitative comparison. Our method successfully reconstructs bright and sharp intensity frames even in complex backgrounds and scenes, demonstrating its robustness on an unrelated open benchmark.}
    \label{fig:EDS}
\end{figure*}

Although there is currently no benchmark for event-based radiance field reconstruction in the dark, we further demonstrated the robustness of our approach through tasks that are not directly related, such as event-based frame reconstruction in the dark. The EDS dataset \cite{hidalgo2022event} provides events and frames under a low-light environment, which is compatible with the input-output format of our approach. However, the dark frames have no paired bright-light ground truth data. The EDS \cite{hidalgo2022event} provides unpaired bright frames captured from a different camera perspective and trajectory. In this case, no quantitative evaluation can be performed since there are no paired data. However, using the bright light frames as a reference, we could qualitatively evaluate our approach to this open benchmark for fairness. Fig. \ref{fig:EDS} demonstrates our rendered results (yellow rows) using the dark frame as input (orange rows) under the peanut and rocket scenes in the EDS \cite{hidalgo2022event}. Sub figures in the red column are bright frames in the same scene, but could only be considered as a qualitative reference since they are not paired with the dark frames. Our approach renders bright and sharp intensity results even in complex backgrounds and scenes. The robust performance of Dark-EvGS on both our proposed dataset and another open benchmark again reflects the effectiveness of our methods.

\subsubsection{Analysis on Challenges and Why Existing Methods Fail}  
Reconstructing real-world radiance fields in low-light and dark environments presents several core challenges that make the task uniquely difficult compared to standard lighting conditions. Below, we analyze the major challenges and discuss why existing methods fail to address them.

\textit{i) Limited Dynamic Range:} Traditional frame-based cameras have a limited dynamic range, making it challenging to capture sufficient details in low-light settings \cite{wan2022learning, zhan2024spiking}. The captured frames often lack critical scene information, resulting in incomplete inputs for radiance field reconstruction. Although event cameras are capable of capturing unseen details in the dark due to their high dynamic range, directly applying frame-based Gaussian Splatting pipelines such as 3D GS~\cite{kerbl20233d} under dark conditions fails, as shown in Fig.~\ref{fig:visual comparison} (c). Without sufficient intensity and brightness cues, these methods fail to capture fine details and produce degraded radiance fields.

\textit{ii) Increased Noise in Event Data:} 
While event cameras are advantageous in low-light scenarios, they inevitably suffer from heightened noise levels under extreme dark conditions \cite{duan2025eventaid}. The increased randomness and sparsity of the event streams make it harder to extract meaningful information (Fig.~\ref{fig:setup} (c)), reducing the reliability of existing event-based radiance field approaches~\cite{rudnev2023eventnerf, wu2024EV-GS, deguchi2024e2gs, wu2025sweepevgs}. The reason lies in physics: like all vision sensors, event cameras are subject to photon-counting noise~\cite{hu2021v2e}. Under low-light conditions, the scarcity of photons causes higher noise and slower event rates, which undermines methods relying purely on events \cite{wu2024EV-GS, rudnev2023eventnerf} or event-assisted radiance field reconstruction \cite{wu2025sweepevgs, deguchi2024e2gs}. These approaches cannot effectively handle the noise induced by dark environments, leading to blurry or unstable reconstructions.

\textit{iii) Uncertain Camera Pose:}  
Accurate pose estimation is another major obstacle in dark environments. On the one hand, structure-from-motion pipelines struggle because dark frames lack sufficient intensity and brightness cues, making it difficult to extract reliable features for geometric matching and tracking. On the other hand, while event cameras provide high temporal resolution, there is currently no mature and robust event-based COLMAP alternative capable of handling noisy and sparse events in such conditions.

In our comparative experiments, for fairness, we supplied all competing methods with ground-truth accurate poses obtained from calibration and known trajectories. However, this setting does not reflect real-world deployment, where such high-quality poses are usually unavailable under dark conditions. In contrast, our pipeline circumvents this difficulty by generating pseudo-bright frames with sufficient structural and photometric cues, enabling COLMAP to recover accurate poses directly without additional steps or specialized tools. Moreover, our dataset contributes beyond benchmarking: it provides precise ground-truth poses alongside real low-light event data, which not only ensures reliable evaluation but also lays the foundation for advancing research toward future \emph{pose-free} dark Gaussian Splatting methods.

\textit{iv) Data Gap Between Synthetic and Real-World Data:}  
Most existing methods, whether event-based radiance field reconstruction~\cite{rudnev2023eventnerf, wu2024EV-GS, wu2025sweepevgs, deguchi2024e2gs} or event-based video enhancement~\cite{rebecq2019events, liang2023coherent}, rely heavily on either synthetic datasets or real-world data with normal light. However, synthetic paired data fail to capture the complexities of real-world low-light conditions, such as diverse noise patterns, dynamic lighting changes, and motion blur. For instance, EvLowLight~\cite{liang2023coherent} is trained on synthetic paired short- and long-exposure frames, along with simulated events~\cite{hu2021v2e}, but these data are too clean and lack realistic noise distributions~\cite{stoffregen2020reducing}. Consequently, when applied to real-world dark frames, EvLowLight \cite{liang2023coherent} fails to enhance inputs effectively and performs poorly when combined with 3D GS for radiance field reconstruction (Fig.~\ref{fig:visual comparison}, column g).  
This highlights the necessity of creating real-world datasets that contrast dark and bright scenes. Our dataset is the first to provide such real-world low-light event data with accurate poses, which not only supports the evaluation of Dark-EvGS but also contributes a valuable benchmark for advancing this research field.

\textit{v) Summary:}  
In summary, low-light radiance field reconstruction suffers from limited dynamic range, increased noise in event data, uncertain camera poses, and a significant gap between synthetic and real-world datasets. These challenges collectively explain why existing methods fail in dark environments. Our Dark-EvGS framework is designed specifically to address these issues, leading to robust reconstruction quality under extreme low-light conditions.

\section{Future Work}
\label{supp: future work}
For future work, we envision expanding Dark-EvGS for broader real-world applications. One promising direction is reconstructing large-scale scenes at night using handheld devices or vehicles. This extension would require optimizing the framework for dynamic scenarios, integrating robust motion compensation techniques, and enhancing the system’s computational efficiency. Another important direction is to address unknown or inaccurate camera poses under dark environments. 
In such scenarios, with severe noise, motion blur, and a lack of visual features, accurate pose information can be unavailable. Future work could explore jointly optimizing camera pose estimation and radiance field reconstruction, potentially leveraging the high temporal resolution and robustness of event data to mitigate pose drift and misalignment in low-light conditions.
This work could pave the way for impactful applications in areas such as autonomous navigation, AR/VR, and nighttime photography by enabling nighttime scene capture and reconstruction in diverse real-world contexts.

\section{Limitation.}
While Dark-EvGS demonstrates significant advancements in radiance field reconstruction under low-light conditions, it is not without limitations. A key challenge is that the current framework only supports static radiance field reconstruction as it was built on the 3D GS pipeline, which does not account for temporal variations effectively. For future work, we will explore dynamic radiance field reconstruction in the dark.

\section{Conclusion}
\label{sec:conclusion}

In this work, we introduce Dark-EvGS, the first event-guided 3D Gaussian Splatting framework for bright frame synthesis from arbitrary views in low-light environments. Leveraging the high dynamic range and speed of event cameras, Dark-EvGS overcomes noise, motion blur, and poor illumination challenges faced by conventional methods. We propose a triplet-level supervision strategy to enhance scene understanding, detail refinement, and sharpness restoration, along with a Color Tone Matching Block to ensure color consistency. Additionally, we introduce the first real-world dataset for event-guided bright frame synthesis in low-light settings. Extensive experiments show that Dark-EvGS outperforms prior approaches, establishing it as a robust solution for low-light radiance field reconstruction.

\section*{Acknowledgments}
This work is supported by the Research Grants Council of Hong Kong (GRF 17201822), the Theme-based Research Scheme (T45-701/22-R), National Natural Science Foundation of China (Grant No. 62402014, 62136001), Beijing Natural Science Foundation (Grant No. L233024), Beijing Municipal Science \& Technology Commission, Administrative Commission of Zhongguancun Science Park (Grant No. Z241100003524012), Taishan Scholars Program (No. TSQN202507241), Shandong Provincial Natural Science Foundation for Young Scholars Program (No. ZR2025QC1627). Peiqi Duan is also supported by China National Postdoctoral Program for Innovative Talents (Grant No. BX20230010) and China Postdoctoral Science Foundation (Grant No. 2023M740076).


\bibliography{TIP-Latex/ref}
\bibliographystyle{IEEEtran}

\end{document}